\documentclass[conference]{IEEEtran}

\makeatletter
 \let\old@ps@headings\ps@headings
 \let\old@ps@IEEEtitlepagestyle\ps@IEEEtitlepagestyle
 \def\confheader#1{%
 \def\ps@headings{%
 \old@ps@headings%
 \def\@oddhead{\strut\hfill#1\hfill\strut}%
 \def\@evenhead{\strut\hfill#1\hfill\strut}%
 }%
 \def\ps@IEEEtitlepagestyle{%
 \old@ps@IEEEtitlepagestyle%
 \def\@oddhead{\strut\hfill#1\hfill\strut}%
 \def\@evenhead{\strut\hfill#1\hfill\strut}%
 }%
 \ps@headings%
 }
 \makeatother

\confheader{}

 \usepackage[pscoord]{eso-pic}
\newcommand{\placetextbox}[3]{
 \setbox0=\hbox{#3}
 \AddToShipoutPictureFG*{ \put(\LenToUnit{#1\paperwidth},\LenToUnit{#2\paperheight}){\vtop{{\null}\makebox[0pt][c]{#3}}}
 }
 }
 \placetextbox{.23}{0.055}{\small{}}

\IEEEoverridecommandlockouts
\usepackage{cite}
\usepackage{amsmath,amssymb,amsfonts}
\usepackage{algorithm}
\usepackage{graphicx}
\usepackage{textcomp}
\usepackage{multirow}
\usepackage{xcolor}
\usepackage{comment}
\usepackage[caption=false]{subfig}
\usepackage{commath}
\usepackage{fancyvrb}
\usepackage{multirow}
\usepackage{array}
\usepackage{algpseudocode}
\usepackage{amsmath}
\usepackage[most]{tcolorbox}
\usepackage{xcolor}
\usepackage{comment}
\usepackage{booktabs}

\usepackage{fancyhdr}
\usepackage{eso-pic}

\def\BibTeX{{\rm B\kern-.05em{\sc i\kern-.025em b}\kern-.08em
    T\kern-.1667em\lower.7ex\hbox{E}\kern-.125emX}}

\title{Evaluating LLM-Generated Rules for Heart Disease Prediction}

\author{
\IEEEauthorblockN{
Feisal Alaswad\textsuperscript{1,*},
Batoul Aljaddouh\textsuperscript{1},
Maher Alrahhal\textsuperscript{2},
Wafaa Al Nassan\textsuperscript{3},
Talal Bonny\textsuperscript{3}
}
\IEEEauthorblockA{
\textsuperscript{1}\textit{Department of Computing Technologies}, 
\textit{SRM Institute of Science and Technology}, Kattankulathur, Chennai, India\\
feisal.alaswad@hotmail.com \qquad batoul.aljaddouh@gmail.com
}

\IEEEauthorblockA{
\textsuperscript{2}\textit{Department of Computer Science}, 
\textit{Amity University Dubai}, Dubai, UAE\\
maherrahal92@gmail.com
}

\IEEEauthorblockA{
\textsuperscript{3}\textit{College of Computing and Informatics}, 
\textit{University of Sharjah},Sharjah, UAE\\
walnassan@sharjah.ac.ae \qquad tbonny@sharjah.ac.ae
}
}

\begin{document}

\maketitle


\confheader{Accepted at The 8th Advances in Science and Engineering Technology (ASET 2026), Dubai, UAE}

\thispagestyle{fancy}
\renewcommand{\headrulewidth}{0pt}

\fancyfoot[C]{%
\parbox{0.95\textwidth}{%
\centering
\footnotesize
\textcopyright\ 2026 IEEE. Personal use of this material is permitted. Permission from IEEE must be obtained for all other uses, in any current or future media, including reprinting/republishing this material for advertising or promotional purposes, creating new collective works, for resale or redistribution to servers or lists, or reuse of any copyrighted component of this work in other works.%
}}


\begin{abstract}
This study compares traditional machine learning models and Large Language Model (LLM)-generated rule-based systems for heart disease prediction using the UCI Heart Disease dataset. Several classifiers, including Logistic Regression, K-Nearest Neighbors (KNN), Support Vector Machine (SVM), Naive Bayes, Decision Tree, and Random Forest, were evaluated alongside rule-based systems generated using GPT-4o and Claude Sonnet 4.6. Model performance was assessed using accuracy, precision, recall, and F1-score metrics. Experimental results show that traditional machine learning models consistently outperform LLM-generated rule-based systems in predictive performance. Random Forest achieved the best overall performance with 90.2\% accuracy, a precision of 0.829, perfect recall of 1.0, and an F1-score of 0.906. Naive Bayes followed closely with 88.5\% accuracy and an F1-score of 0.881. In contrast, the LLM-generated rule models achieved lower performance, with Claude Sonnet 4.6 reaching 80.3\% accuracy (F1-score: 0.833) and GPT-4o obtaining 70.5\% accuracy (F1-score: 0.690). Despite the performance gap, the LLM-generated rules provide interpretable IF–THEN diagnostic logic that enhances explainability and transparency in clinical decision-making. These findings highlight the trade-off between predictive performance and interpretability in medical artificial intelligence systems. The complete implementation of all experiments, including machine learning models and LLM-derived rule classifiers, is publicly available in the GitHub repository at https://github.com/FeisalAlaswad/LLM-Rule-ML-Heart-Disease-Prediction .
\end{abstract}

\begin{IEEEkeywords}
Heart disease prediction, machine learning, large language models, rule extraction, interpretable artificial intelligence, clinical decision support, Random Forest, healthcare analytics
\end{IEEEkeywords}

\section{Introduction}
Artificial intelligence has increasingly been investigated as a complementary approach to conventional cardiovascular risk assessment because ML models can integrate multiple heterogeneous clinical variables and identify complex associations that may be difficult to capture using conventional statistical approaches~\cite{topol2019high,liu2023machine}. However, improvements in predictive performance alone are insufficient for clinical adoption, where transparency, reliability, and appropriate human oversight are also important considerations~\cite{topol2019high,rudin2019stop}.

Heart disease is one of the leading causes of death worldwide and remains a major public health challenge~\cite{who_cvd}. Early prediction and diagnosis of cardiovascular disease are essential for reducing mortality rates and improving patient outcomes~\cite{el2024proposed,banerjee2025systematic}. Clinical diagnosis often depends on multiple physiological and medical factors such as blood pressure, cholesterol level, chest pain characteristics, electrocardiographic measurements, and exercise-related symptoms~\cite{khan2024comprehensive,hasan2026advancing}. However, analyzing these complex relationships manually can be difficult and time-consuming~\cite{pal2022risk}.

Machine learning (ML) and deep learning (DL) techniques have been widely adopted in medical diagnosis because of their ability to learn patterns from clinical datasets and generate accurate predictive models~\cite{asif2025advancements,aljaddouh2024multimodal,alrahhal2025enhancing,alanazi2017critical,aljaddouh2026cxl,aljaddouh2023boundary}. Traditional supervised learning algorithms such as Logistic Regression, K-Nearest Neighbors (KNN), Support Vector Machines (SVM), Naive Bayes, Decision Trees, and Random Forests have demonstrated strong performance~\cite{khachane2017organ,alaswad2021categorising,caruana2006empirical,nasir2021hypertension}. These models can identify nonlinear relationships among patient attributes and provide reliable predictions for disease detection~\cite{dinh2019data,talaat2025toward,xu2025non}. Comparative studies have reported that the relative performance of ML algorithms can vary substantially across cardiovascular datasets, patient populations, and evaluation settings, indicating that no single classifier consistently provides the best performance across all clinical prediction tasks~\cite{liu2023machine}. This variability highlights the importance of evaluating multiple learning algorithms rather than relying on a single predictive model when developing data-driven cardiovascular decision-support systems.

Despite their predictive effectiveness, many machine learning models operate as black-box systems with limited interpretability~\cite{bodria2023benchmarking,hassija2024interpreting}. In medical applications, interpretability is particularly important because clinicians require understandable reasoning behind automated predictions~\cite{alrahhal2025explainable,hatherley2024virtues,alrahhal2025xai,aljaddouh2026multimodal}. Rule-based systems and decision trees are commonly used to improve transparency by representing predictions using human-readable IF--THEN decision rules~\cite{kostopoulos2024explainable,alaswad2023software}.

Recently, Large Language Models (LLMs) have emerged as powerful artificial intelligence systems capable of reasoning over structured and unstructured information~\cite{liu2024suql,alaswad2026cocomo,kojima2022large,jayaseeli2026integrating,alaswad2026hybrid}. Beyond natural language generation, LLMs have shown the ability to generate interpretable diagnostic rules and clinical decision logic from datasets or prompts~\cite{sivasothy2026large,savage2024diagnostic}. This introduces the possibility of using LLM-generated rules as explainable alternatives to conventional machine learning approaches~\cite{balek2025llm,longo2025computational,koebler2024more}.

This study compares traditional machine learning classifiers with rule-based systems generated using LLMs for heart disease prediction. Several machine learning models are evaluated alongside rules extracted by LLMs. Experimental results show that traditional machine learning models outperform LLM-generated rule systems in predictive performance, while LLM rules provide more interpretable and human-readable diagnostic logic.

The remainder of this paper is organized as follows. Section~\ref{sec:relatedwork} reviews previous studies on machine learning-based heart disease prediction, interpretable rule-based approaches, and the emerging use of large language models for clinical decision support. Section~\ref{sec:methodology} describes the dataset, preprocessing procedures, machine learning models, LLM-based rule extraction process, and experimental workflow. Section~\ref{sec:results} presents and discusses the experimental results, including the comparative performance of conventional machine learning models and LLM-generated rule-based classifiers. Finally, Section~\ref{sec:conclusion} concludes the paper by summarizing the main findings, discussing the trade-off between predictive performance and interpretability, and outlining directions for future research.

\section{Related Work}\label{sec:relatedwork}
Heart disease prediction has been widely studied using traditional machine learning techniques due to the increasing availability of clinical datasets and the need for early diagnostic support systems~\cite{maini2021machine,ahsan2022machine}. Classical machine learning models such as Logistic Regression, KNN, SVM, Decision Trees, Random Forests, and Naive Bayes have demonstrated strong performance in predicting cardiovascular diseases from structured clinical attributes including chest pain type, cholesterol level, blood pressure, age, and electrocardiographic measurements~\cite{rahim2021integrated,ahmed2024heart,iacobescu2024evaluating,noroozi2023analyzing}. Among these methods, ensemble approaches such as Random Forest often achieve high predictive accuracy because of their ability to capture nonlinear relationships and feature interactions~\cite{ramalingam2018heart,yahaya2020comprehensive}.

In addition to predictive performance, interpretability has become an important research direction in medical artificial intelligence. Decision Trees and rule-based systems are frequently used because they provide human-readable IF–THEN rules that can support clinical understanding and transparency. Several studies have focused on extracting interpretable rules from trained machine learning models to improve trustworthiness and explainability in healthcare applications~\cite{chakraborty2024rule,zhang2023predicting}.

Beyond text generation, LLMs have shown potential for generating diagnostic rules, clinical explanations, and decision-support logic from datasets or prompts~\cite{liu2024suql,sandmann2024systematic}. Models such as GPT-4o and Claude Sonnet can synthesize interpretable IF–THEN decision rules that resemble expert reasoning. This has motivated growing interest in evaluating whether LLM-generated rules can serve as lightweight and explainable alternatives to conventional machine learning approaches.

However, despite the increasing use of LLMs in healthcare reasoning tasks~\cite{berger2025reasoning,mansoor2025reasoning,peng2026aligning}, limited work has directly compared manually prompted LLM-extracted rules against traditional supervised learning models on structured heart disease datasets. Furthermore, there remains insufficient analysis regarding the trade-off between interpretability and predictive performance when comparing symbolic rules generated by LLMs to data-driven classifiers.

\section{Methodology}
\label{sec:methodology}

\begin{figure*}[!t]
    \centering
    \includegraphics[width=0.95\textwidth]{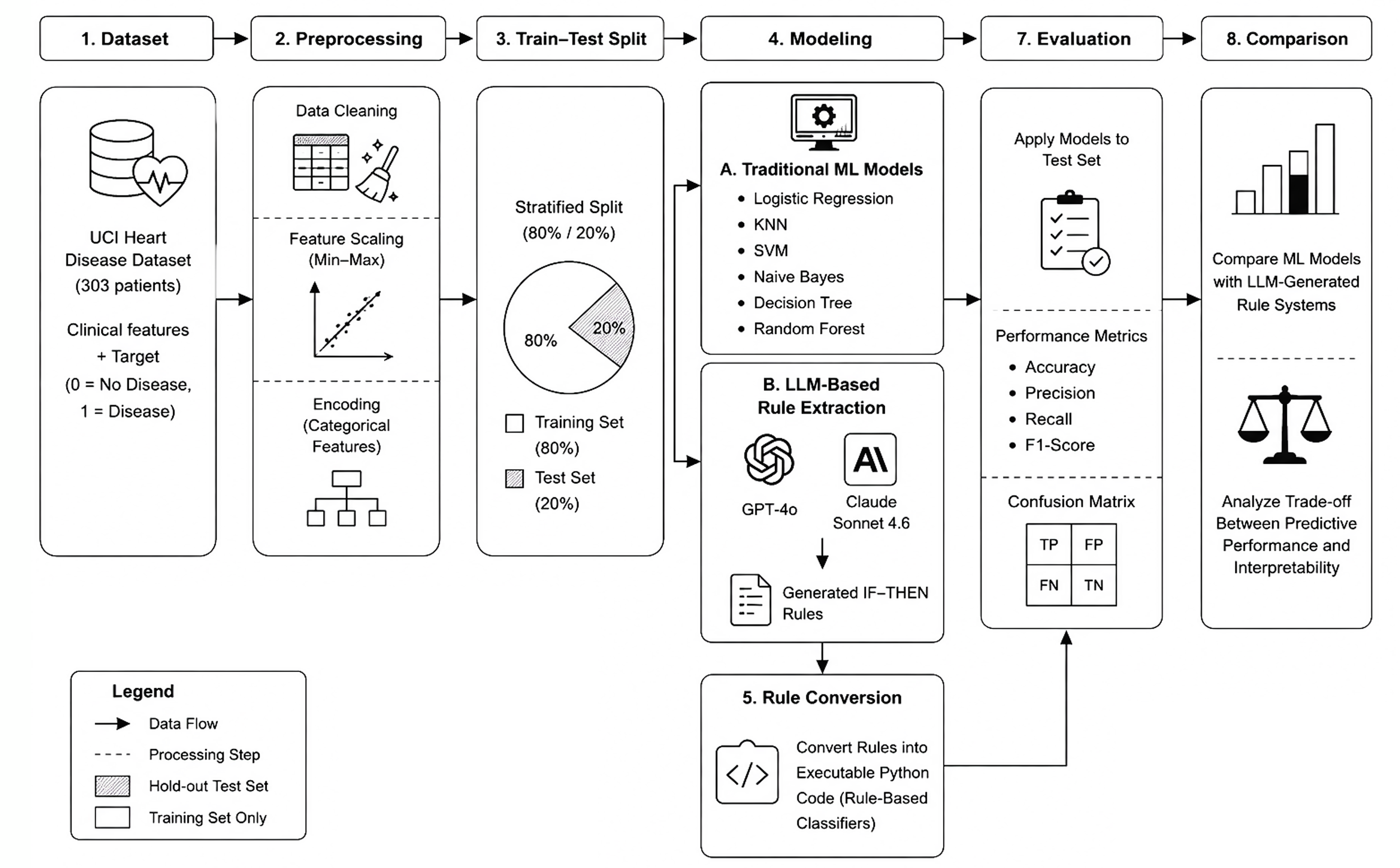}
    \caption{Overall experimental workflow for comparing traditional machine learning classifiers with LLM-generated rule-based classifiers for heart disease prediction.}
    \label{fig:workflow}
\end{figure*}

This study investigates the effectiveness of LLM-generated clinical rules for heart disease prediction and compares their performance against traditional machine learning classifiers. The overall workflow consists of dataset preprocessing, model training, rule extraction using LLMs, and performance evaluation. The overall experimental workflow is illustrated in Fig.~\ref{fig:workflow}. The pipeline begins with preprocessing the UCI Heart Disease dataset and performing a stratified train--test split, followed by two parallel modeling approaches: conventional machine learning classifiers and LLM-based rule extraction using GPT-4o and Claude Sonnet 4.6. The generated rules are subsequently converted into executable rule-based classifiers and evaluated on the same held-out test set using standard classification metrics before the predictive performance of the two approaches is compared.

\subsection{Dataset Description}

The experiments were conducted using the publicly available UCI Heart Disease dataset, which contains clinical records collected from 303 patients. The dataset includes demographic information, physiological measurements, exercise test results, and cardiac examination attributes commonly associated with coronary artery disease~\cite{uci_heart_disease_45}.

Each patient record is represented using several clinical features, including age, sex, chest pain type, fasting blood sugar, resting electrocardiographic results, exercise-induced angina, ST segment slope, thalassemia test outcomes, number of major vessels observed through fluoroscopy, resting blood pressure, serum cholesterol, maximum heart rate achieved, and exercise-induced ST depression.

The target variable represents the presence or absence of heart disease. In this study, the problem is formulated as a binary classification task, where 0 indicates no heart disease and 1 indicates the presence of heart disease.

\subsection{Data Preprocessing}

Prior to training, the dataset was cleaned and prepared for machine learning analysis. The target attribute was separated from the input features, and all numerical attributes were normalized using Min-Max scaling to ensure that feature values were mapped into the range $[0,1]$. Feature scaling was applied to improve the convergence and stability of distance-based and gradient-based classifiers.

The dataset was divided into training and testing subsets using an 80:20 split. Stratified sampling was employed to preserve the original class distribution in both subsets.

\subsection{Machine Learning Models}

Several supervised machine learning classifiers were implemented to establish baseline predictive performance, including Logistic Regression, KNN, SVM, Gaussian Naive Bayes, Decision Tree, and Random Forest. Each classifier was trained using the training subset and evaluated on the unseen testing subset. All models were implemented using the Scikit-learn library in Python.

\subsection{LLM-Based Rule Extraction}

In addition to conventional machine learning models, this study explores the ability of large language models (LLMs) to generate interpretable clinical decision rules for heart disease prediction. Specifically, two LLM systems—GPT-4o and Claude Sonnet 4.6—were employed to derive rule-based classifiers.

Only the training subset of the UCI Heart Disease dataset was made available to the language models for rule generation. The held-out test instances and their corresponding target labels were not provided to either GPT-4o or Claude Sonnet 4.6 at any stage of rule generation. The models therefore generated the IF--THEN rules exclusively from the training-data context, while the independent test set was reserved solely for the final evaluation of the resulting fixed rule sets.

The generated rules were manually converted into executable rule-based classifiers using Python conditional statements. Each rule-based system produced binary predictions directly from patient attributes without additional training.

\subsection{Performance Evaluation}

The predictive performance of all machine learning models and LLM-generated rule systems was evaluated using standard classification metrics, including accuracy, precision, recall, and F1-score. These metrics were computed using the test dataset to assess both predictive capability and classification reliability.

Comparative analysis was then performed to determine how the interpretable LLM-generated rules compare against traditional data-driven machine learning approaches.

\subsection{Experimental Workflow}

The complete experimental pipeline followed the steps below:

\begin{enumerate}
    \item Load and preprocess the UCI Heart Disease dataset.
    \item Normalize feature values using Min-Max scaling.
    \item Split the dataset into training and testing subsets.
    \item Train conventional machine learning classifiers.
    \item Generate clinical IF--THEN rules using LLMs.
    \item Convert generated rules into executable rule-based classifiers.
    \item Evaluate all approaches using classification metrics and confusion matrices.
    \item Compare predictive performance between machine learning and LLM-generated rule systems.
\end{enumerate}

\subsection{LLM-Based Rules Extracted}

To investigate the interpretability of LLMs in medical decision-making, we prompted GPT-4o to generate explicit IF--THEN decision rules based on the UCI dataset features. The resulting rule-based classifier was implemented as a deterministic function, enabling direct comparison with traditional machine learning models under identical evaluation settings.

The extracted GPT-4o and Claude Sonnet 4.6 rule sets are presented in Algorithm~\ref{alg:gpt4o_rules} and Algorithm~\ref{alg:sonnet_rules}, respectively.

\begin{algorithm}[h]
\caption{GPT-4o Rule-Based Classifier}
\label{alg:gpt4o_rules}
\begin{algorithmic}[1]

\Require Feature vector $X = \{age, cp, \ldots\}$
\Ensure Predicted class $y \in \{0,1\}$

\If{$cp > 0 \land thal \leq 2 \land thalach > 150$} \Return $1$
\ElsIf{$cp > 0 \land thal \leq 2$} \Return $1$
\ElsIf{$cp = 0 \land ca > 0$} \Return $0$
\ElsIf{$cp = 0 \land thal > 2$} \Return $0$
\ElsIf{$cp = 0 \land exang = 1$} \Return $0$
\ElsIf{$oldpeak > 2$} \Return $0$
\ElsIf{$age < 56 \land cp > 0$} \Return $1$
\Else \ \Return $0$
\EndIf

\end{algorithmic}
\end{algorithm}

\begin{algorithm}[h]
\caption{Claude Sonnet 4.6 Rule-Based Classifier}
\label{alg:sonnet_rules}
\begin{algorithmic}[1]

\Require Feature vector $X = \{age, cp, \ldots\}$
\Ensure Predicted class $y \in \{0,1\}$


\If{$cp > 0 \land thal = 2 \land oldpeak \leq 3.5$} \Return $1$
\ElsIf{$cp > 0 \land thal = 3 \land thalach > 150 \land oldpeak \leq 2.2$} \Return $1$
\ElsIf{$cp = 0 \land oldpeak \leq 0.7 \land ca = 0 \land age > 41$}  \Return $1$
\ElsIf{$cp = 0 \land ca > 0 \land sex = 0$} \Return $1$


\ElsIf{$cp = 0 \land oldpeak > 0.7 \land trestbps > 106$} \Return $0$
\ElsIf{$cp > 0 \land thal = 3 \land thalach \leq 150$} \Return $0$
\ElsIf{$cp = 0 \land ca > 0 \land sex = 1$} \Return $0$
\ElsIf{$oldpeak > 3.5$} \Return $0$


\ElsIf{$cp > 0$} \Return $1$
\Else \ \Return $0$
\EndIf

\end{algorithmic}
\end{algorithm}

The extracted rule sets show that both GPT-4o and Claude Sonnet 4.6 primarily rely on clinically meaningful features such as chest pain type ($cp$), thalassemia status ($thal$), number of major vessels ($ca$), exercise-induced angina ($exang$), ST depression ($oldpeak$), and related cardiovascular indicators. While these rules are highly interpretable and align with medical intuition, they remain heuristic in nature and do not incorporate probabilistic calibration or robust feature interaction learning. This limitation, particularly in GPT-4o’s simpler rule structure compared to the more structured Sonnet 4.6 rules, contributes to their lower predictive performance relative to traditional supervised learning models.

\subsection{LLM Prompting and Reproducibility}

To ensure reproducibility, the LLM-based rule extraction was performed using GPT-4o (model identifier: \texttt{gpt-4o}) and Claude Sonnet 4.6 (model identifier: \texttt{claude-sonnet-4-6}). The default generation settings were used, with temperature, top-$p$, and maximum output length left at their respective API defaults. Each model was prompted once using the same prompt and dataset description. No iterative prompting, response refinement, or model-specific rule correction was performed.

The prompt provided the clinical feature definitions, binary prediction objective, and training-data context, and explicitly requested a compact set of human-readable IF--THEN rules for predicting heart disease. The same semantic prompt was provided to both models to ensure a consistent comparison:

\begin{quote}
You are an expert in machine learning and clinical decision-support systems. Using the UCI Heart Disease dataset and the feature definitions provided below, derive a concise and interpretable rule-based classifier for binary heart disease prediction.

The target variable is binary: 0 indicates no heart disease and 1 indicates the presence of heart disease. Use only the clinical features provided in the dataset. Identify meaningful combinations of features and formulate explicit IF--THEN rules. Each rule must contain clearly defined feature conditions and must return either 0 or 1.

The rules should be deterministic, human-readable, and suitable for direct implementation as Python conditional statements. Prefer clinically meaningful feature combinations, avoid introducing external medical knowledge or unsupported assumptions, and provide a default rule for cases that are not covered by the preceding conditions.

Do not train a machine-learning model, perform statistical optimization, or use information from external datasets. Do not estimate probabilities. Return the final decision rules together with the feature conditions and predicted class in an explicit, executable form.
\end{quote}

The original responses generated by GPT-4o and Claude Sonnet 4.6 were retained before conversion and were used as the source for the corresponding rule sets. Manual conversion was performed by translating each explicitly stated IF--THEN condition into an equivalent Python conditional statement while preserving the original feature names, threshold values, logical operators, rule ordering, and predicted class. No additional rules or thresholds were introduced during the conversion process. Explanatory text generated by the models was excluded from execution and was not interpreted as an additional rule. The resulting deterministic classifiers were then applied to the same held-out test set used for evaluating the conventional machine learning models. This procedure ensured that the comparison reflected the predictive performance of the generated rules rather than subsequent human optimization of the LLM outputs.

\textbf{Training--Test Separation and Information Leakage Prevention.} LLM rule generation was restricted to the training set only. Neither GPT-4o nor Claude Sonnet 4.6 received the held-out test instances, labels, predictions, or performance results. The generated rules were frozen before being applied to the unseen test set, ensuring that test data were used exclusively for final evaluation and that no label leakage or test-driven rule construction occurred.

\section{Results and Discussion}
\label{sec:results}

This section presents the experimental results obtained from both traditional machine learning models and LLM-generated rule-based classifiers for heart disease prediction. The models were evaluated using Accuracy, Precision, Recall, and F1-Score.

\subsection{Performance Comparison}

\begin{figure*}[!htbp]
    \centering
    \includegraphics[width=0.9\linewidth]{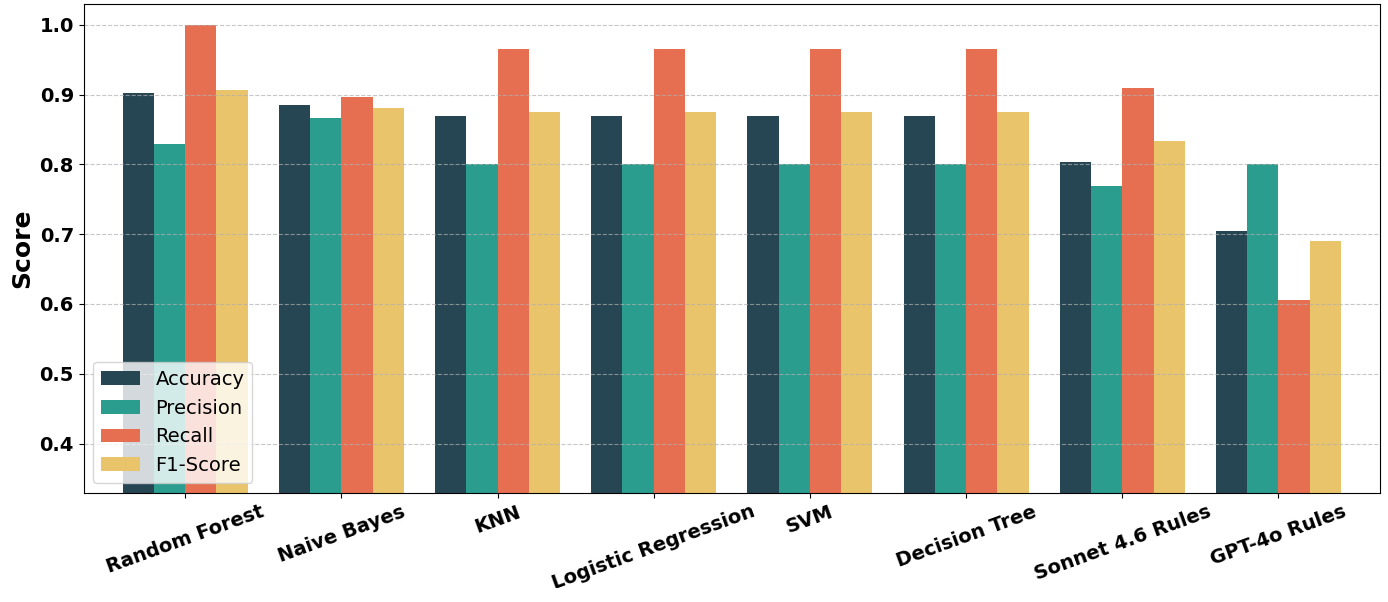}
    \caption{Performance comparison between ML models and LLM-generated rule-based classifiers}
    \label{fig:performance}
\end{figure*}

Table~\ref{tab:results} summarizes the performance of all evaluated approaches.  Several classical models (KNN, Logistic Regression, SVM, and Decision Tree) achieved identical scores due to the small, highly separable dataset, where different algorithms converge to similar decision boundaries and produce nearly identical predictions. Figure~\ref{fig:performance} illustrates the comparative performance of all models across the evaluation metrics.

\begin{table}[h]
\centering
\caption{Performance comparison of ML and LLM-based rule models}
\label{tab:results}
\begin{tabular}{lcccc}
\hline
\textbf{Model} & \textbf{Accuracy} & \textbf{Precision} & \textbf{Recall} & \textbf{F1-Score} \\
\hline
Random Forest & 0.902 & 0.829 & 1.000 & 0.906 \\
Naive Bayes & 0.885 & 0.867 & 0.897 & 0.881 \\
KNN & 0.869 & 0.800 & 0.966 & 0.875 \\
Logistic Regression & 0.869 & 0.800 & 0.966 & 0.875 \\
SVM & 0.869 & 0.800 & 0.966 & 0.875 \\
Decision Tree & 0.869 & 0.800 & 0.966 & 0.875 \\
Sonnet 4.6 Rules & 0.803 & 0.769 & 0.909 & 0.833 \\
GPT-4o Rules & 0.705 & 0.800 & 0.606 & 0.690 \\
\hline
\end{tabular}
\end{table}

\subsection{Discussion}
The experimental results demonstrate that traditional machine learning models significantly outperform the rule-based classifiers generated using LLMs. Among all evaluated approaches, the Random Forest classifier achieved the best overall performance, with an accuracy of 90.2\% and an F1-score of 0.906. The model also achieved perfect recall (1.0), indicating that all positive heart disease cases in the test set were correctly identified.

Naive Bayes achieved the second-best performance with an accuracy of 88.5\%, an F1-score of 0.881, and a balanced trade-off between precision (0.867) and recall (0.897). KNN, Logistic Regression, SVM, and Decision Tree all achieved identical performance, each reaching 86.9\% accuracy and an F1-score of 0.875, with relatively high recall (0.966) but moderate precision (0.800). These results indicate that conventional supervised learning methods are effective in capturing patterns in the clinical dataset, particularly in terms of sensitivity toward positive cases.

In contrast, the LLM-generated rule-based systems show noticeably lower predictive performance. The Sonnet 4.6 rule set achieved an accuracy of 80.3\% and an F1-score of 0.833, with relatively high recall (0.909) but lower precision (0.769), indicating a tendency to over-predict positive cases. The GPT-4o rule set performed worse overall, achieving 70.5\% accuracy and an F1-score of 0.690. Although it maintained moderate precision (0.800), its recall dropped to 0.606, suggesting that a substantial number of positive heart disease cases were missed. Overall, these results highlight a clear performance gap between conventional machine learning models and LLM-generated rule-based systems on this dataset.

These findings suggest that LLMs can generate human-readable IF--THEN rules; however, the clinical validity and usefulness of these rules were not formally evaluated in this study. On the evaluated dataset, the extracted rules did not match the predictive performance of the data-driven machine learning algorithms. Traditional models benefit from statistical optimization and the ability to learn complex feature interactions directly from the data, whereas manually interpreted rule systems are generally constrained by simplified decision boundaries. In particular, both LLM-based rule sets (Sonnet 4.6 and GPT-4o) underperformed all classical machine learning models in terms of F1-score, indicating a weaker balance between precision and recall on this dataset.

Nevertheless, the LLM-generated rules may offer greater transparency because their IF--THEN logic is human-readable and easier to inspect than many conventional predictive models. However, interpretability and clinical validity were not quantitatively evaluated in this study; therefore, these potential advantages should not be interpreted as evidence of improved clinical usability or decision-making. Similarly, the observed difference between predictive performance and rule transparency is better described as a potential accuracy--interpretability trade-off rather than a formally established trade-off. The results should therefore be viewed as an initial comparison of predictive performance and rule transparency rather than evidence supporting clinical deployment or clinical decision support.

\section{Conclusion}\label{sec:conclusion}

This study compared traditional machine learning models with LLM-generated rule-based systems for heart disease prediction using the UCI Heart Disease dataset. The experimental results showed that conventional machine learning models achieved higher predictive performance than the LLM-generated rule sets, with Random Forest obtaining the highest accuracy and F1-score among the evaluated approaches.

The LLM-generated rules produced human-readable IF--THEN logic, suggesting potential benefits for transparency and inspection of model reasoning. However, interpretability and clinical validity were not quantitatively assessed, and the relatively small dataset limits the extent to which these findings can be generalized to real-world clinical populations. Consequently, the results should not be interpreted as evidence that the LLM-based rules are suitable for clinical decision support or deployment. Rather, they provide an exploratory comparison between data-driven predictive models and LLM-generated rule-based approaches.

Future work should evaluate these approaches on larger and more diverse clinical datasets and incorporate formal assessments of interpretability, clinical validity, robustness, and generalizability. Hybrid approaches that combine the predictive capabilities of machine learning with explicitly validated and human-interpretable reasoning may also be investigated.

\bibliographystyle{IEEEtran} 
 
\bibliography{sn-bibliography}

\end{document}